\documentclass[letterpaper]{article}
\usepackage{aaai}
\usepackage{times}
\usepackage{helvet}
\usepackage{courier}
\usepackage{graphicx}
\usepackage{amsmath}
\begin{document}
%
\title{FakeSafe: Human Level Data Protection by Disinformation Mapping using Cycle-consistent Adversarial Network}
\author{He Zhu,Dianbo Liu\\
Department of Biomedical Informatics, Harvard University\\
Computer Science and artificial Intelligence Laboratory, MIT\\
The Hong Kong Polytechnic University\\
Corresponds to: dianbo@mit.edu
}
\maketitle

\section{Introduction}

One of the best way for data protection is to hide real information in fake information. The English term "disinformation" is defined as "false information with the intention to deceive opinion" with slightly negative meaning. However, the idea of disinformation can be borrowed into data science world for private data protection. As more and more data are becoming digitized, efficiency of data transfer, replication and usage has increased significantly in recent years. In addition, due to the development of system like federated data networks, data sharing among large number of institutions and devices becomes possible \cite{jordon2018pate,liu2019two,mukherjee2019protecting,shao2019privacy}. These advances bring convenience to our society. Nevertheless, they also raise big concerns on data security and privacy, especially in sensitive fields like healthcare and personal finance. Despite efforts in data encryption, secure computation and other methods trying to protect data, there is always a chance that data can be leaked due to either technological or human reasons. In addition to technological level protection of data, human level protection is often overlooked \cite{jensen2013challenges,lee2009ethical,joly2016data}. In this study, we aim at developing a technology that provides an additional layer of protection of private data
by mapping the original information onto a fake domain that looks realistic but unidentifiable to human. By doing this, even if the data is leaked, the malicious attackers will not be able to know whether the data they obtained are fake, and, therefore, unable to retrieve the original real information, which is similar to the concept of disinformation.

\par
One of the biggest promises of deep learning is its ability to discover rich representation and approximate complex mapping functions \cite{goodfellow2014generative,bengio2009learning}. Recent advances in generative models such as generative adversarial networks (GANs) utilized these properties of deep learning and significantly simplified some previously difficult tasks such as image-to-image translation and images based text generation \cite{zhu2017unpaired,mirza2014conditional}. Derivatives and modifications of GAN based models have been used and achieved outstanding performance in many fields \cite{odena2017conditional,makhzani2015adversarial,radford2015unsupervised}. These tools can also be used to protect data privacy.

\par 
GANs and its derivatives have achieved impressive result in data generation, style transferring and many other fields \cite{goodfellow2014generative,zhu2017unpaired}. The ability of GAN related models to generate data that are indistinguishable from the real data comes from the idea of adversarial loss. Cycle consistency is a concept originated from machine translation, where a phrase translated from one language to another, after translated back, should be identical to the original phrase. Cycle consistency has been widely applied in machine learning especially in computer vision related tasks \cite{godard2017unsupervised,zhou2016learning}. One recent success in combing GANs and cycle consistency was in image style transferring by CycleGAN \cite{zhu2017unpaired}. In this study, we combined GANs and consistency loss to develop a method named FakeSafe to map the private information onto a fake message that looks indistinguishable from the real messages. The fake message can be either from the same domain of the original private information, or from a completely different domain. FakeSafe can be used during data transfer, data storage, data usage or other scenarios in combination of traditional encryption and security technologies. Using toy data sets as well as real world clinical data set, we conducted a proof-of-concept experiments to explore how well FakeSafe can help protect private information at human level and the quality of reconstructed data from fake domain.

\section{FakeSafe}
\subsection{Motivation and formulation}
The purpose of FakeSafe method is to map the original private information onto a fake but realistically looking message. The method consists of two parts: 1) a function $F$ that maps a private message $X$ into a fake message  $X^{fake}$ , i.e. $X^{fake}=F(X)$. $X^{fake}$ can be from the same domain as $X$, such as a human face image, or a completely different domain. 2) a reconstruction function $R$ that maps the fake message back to original message. $F$ and $R$ are specific to each data set. 

\subsection{System}
In our system of interest, we are assuming there is a sender of private information, a targeted receiver of information. The data transfer/storage infrastructure is not 100\% safe and the malicious attackers might try to steal the private information. Only the data sender has access to function $F$ to map the private data to fake domain. Only targeted receiver has access to function $R$ to retrieve the original data. Even if the attacker obtains $X^{fake}$, without additional information, it is impossible for him or her to know that the data is fake, because $X^{fake}$ looks realistic. If $X$ and $X^{fake}$ are from the same data domain, e.g. mapping a set of human faces into another set of human faces, even if the attacker knows before hand what the data is supposed to look like, it is difficult to notice if the data has been mapped by FakeSafe method into fake messages.

\begin{figure}
  \centering
  {\setlength{\fboxsep}{0pt}%
  \setlength{\fboxrule}{0pt}%
  \fbox{\includegraphics[width=8.25cm, height=13.2cm]{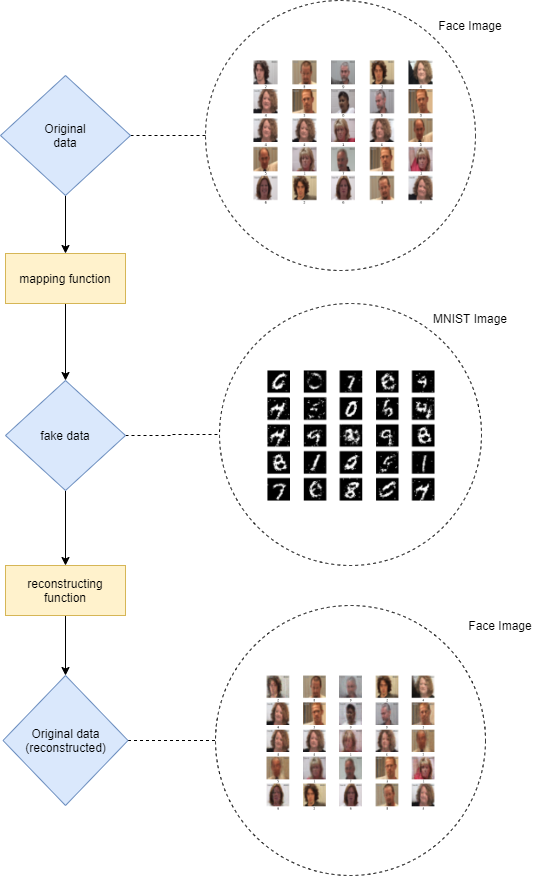}}
}%
 \label{Demo}
  \caption{Human level data protection by mapping original information into fake data domain (disinformation). The original information can be recovered from fake messaged using a trained reconstruction function}
\end{figure}

\begin{figure*}
  \centering
  {\setlength{\fboxsep}{0pt}%
  \setlength{\fboxrule}{0pt}%
  \fbox{\includegraphics[width=12cm, height=8cm]{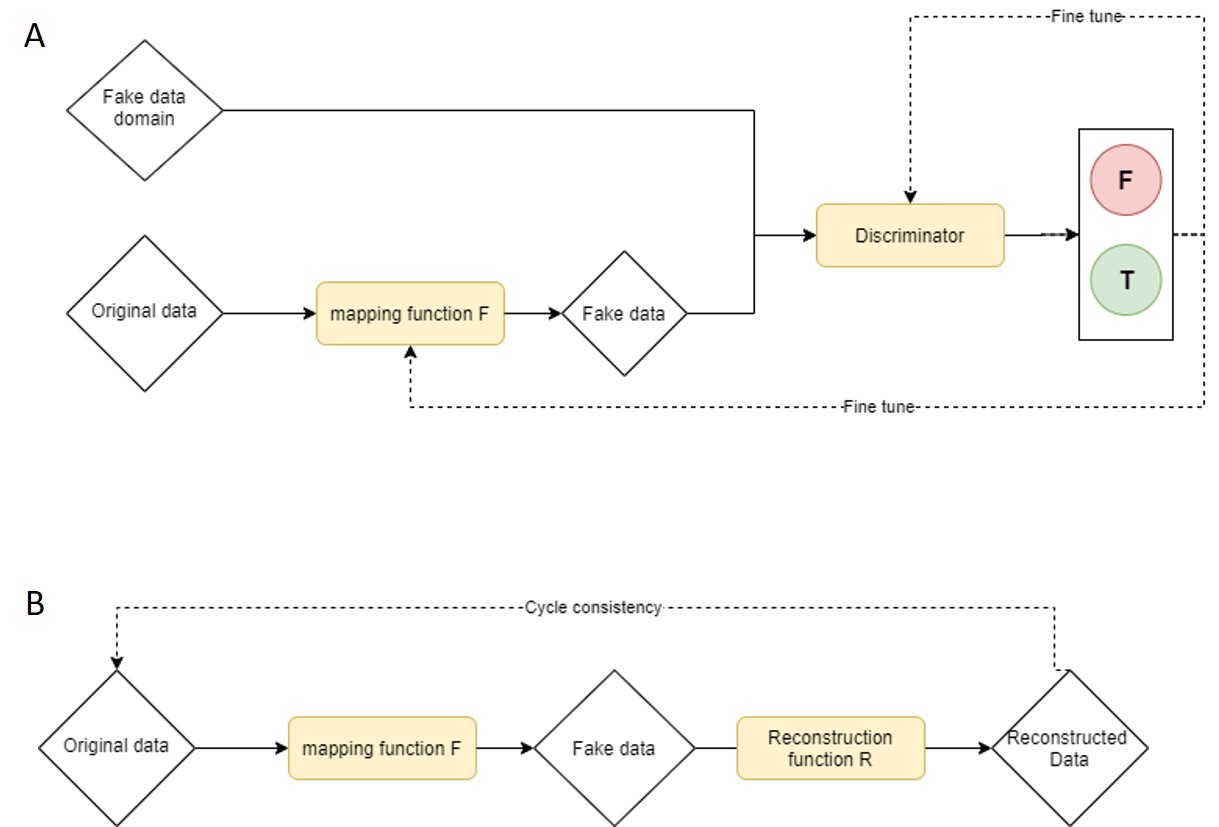}}
}%
 \label{Mechanism}
  \caption{FakeSafe method uses a Generative Adversarial Network (GAN) with cycle consistency to map original private data to fake data (A) and reconstruct original information from the fake message (B) }
\end{figure*}

\subsection{Generative adversarial networks(GAN) with cycle consistency loss}
GAN was used as the function $F$ to map private information to fake message in this study due its good performance in generating fake data sets that look realistic to human. Generative model $F(X)$ is trained against discriminator $D$ to make the outputs of $X^{fake}=F(X)$ look indistinguishable from the samples used to train $D$. We name this type of $X^{fake}$ messages as FakeSafe messages. $D$ and $F$ were trained in an alternating manner. The objective loss function for training generator and discriminator is:

\begin{multline}
L(x,x^{fake},F,D)=\\E[log(D(X^{fake}))]+E[log(1-D(F(X)))]
\end{multline}

$F$ generates data points that look indistinguishable from real data in fake message domain. Least loss was used to train the GAN due to reported stability \cite{zhu2017unpaired}. Therefore, when training GAN, we train $F$ to minimize $E [1-D(F(X))]^2$ and train $D$ to minimize $E [D(F(X))^2]$. 
After the other party receives the FakeSafe message, it will be recovered using a trained model $R$ such that $R(F(X))\approx X$. To enable the ability of $R$ to retrieve the original message from $X^{fake}$, cycle-consistency was used to make reconstructed data $R(X^{Fake})$ matching the original data $X$. The loss function is 

\[L(F,R,X)=E[||R(F(X))-X||]\]

 For reconstruction errors, we used absolute loss. For simplicity, fully connected neural network with leaky ReLU was used in both generator and discriminator models.
 
\subsection{Model implementation}
As this is a proof-of-concept study, 1) for the image-image generator model, we used a simple 3 layer fully connected neural network with 256, 512 and 1024 units. 2) For the text-image generator model, we used a 4 layer fully connected neural network with 64, 256, 512 and 1024 units. Leaky ReLU was used as the activation functions for hidden layers and batch normalization was applied in both image-image and text-image generator model. 3) For the image-text generator model, we used a 4 layer fully connected neural network with 128, 256, 512 and 1024 units. Leaky ReLU was also used as the activation function, and a dropout with rate 0.2 was introduced to avoid over-fitting. The adam optimization with learning rate of 0.0002 was used in the above 3 cases.

\section{Experiments and results}
In order to understand whether our FakeSafe method works in protecting information transfer, we conducted three types of proof-of-concept experiments. First, we encoded information into fake messages from the same data domain, using MNIST and MNIST fashion as example. Second, we encoded information into fake messages from a different domain, such as MNIST digitals to MNIST fashion. Last, we explored the possibility of multi-step FakeSafe encoding of information. The reconstructing accuracy decreased as we increased the number of steps to encode information.
In addition, we tested its potential values in real world application using a face video frame from a clinical settings using an open source data set.  

\begin{figure}
  \centering
  {\setlength{\fboxsep}{0pt}%
  \setlength{\fboxrule}{0pt}%
  \fbox{\includegraphics[width=8cm, height=4cm]{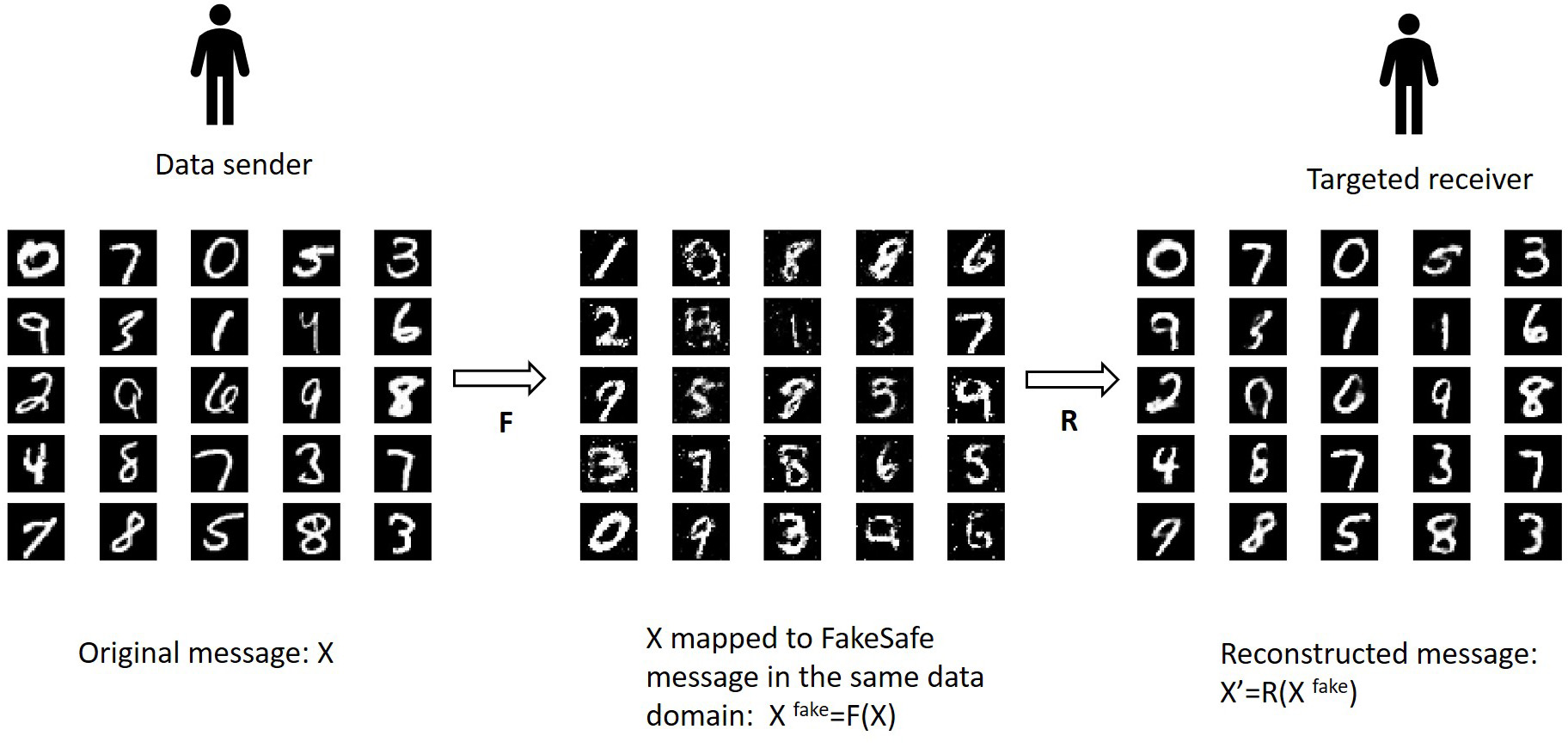}}
}%
\label{SameDomain}
\caption{Use FakeSafe method to map private information into fake data from the same domain }
\end{figure}

\subsection{Date set}
To conduct proof-of-concept experiments, four data sets were used in this study:
1) MNIST hand written digits data set, 2) MNIST fashion data set, 3) an English text data set from Tatoeba  4) the UNBC-McMaster Shoulder Pain Expression Archive Data set \cite{lucey2011painful}. UNBC-McMaster Shoulder Pain is a real world data set from clinical setting and consists of human face video frames from different individuals with shoulder pain. We use this data set as an example of real use case of FakeSafe in medical setting. 

\subsection{FakeSafe mapping onto the same data domain}
One potential application of FakeSafe is to map private information on other same data domain but different data points. We conducted four experiments: MNIST$\rightarrow$F$\rightarrow$MNIST$\rightarrow$R$\rightarrow$MNIST, Fashion$\rightarrow$F$\rightarrow$Fashion$\rightarrow$R$\rightarrow$Fashion, Face$\rightarrow$F$\rightarrow$Face$\rightarrow$R$\rightarrow$Face and Word$\rightarrow$F$\rightarrow$Word$\rightarrow$R$\rightarrow$Word.

\par
When conducting experiments on MNIST, Models $F,D,R$ were all trained using the training sets with images of 10 hand-written digits. Therefore, $X^{fake}=F(X)$ can be any possible number from the training set and might not have to be the same digits as $X$. As shown in figure \ref{SameDomain} , the recovered images $R(F(X))$ have the same labels as the original images $X$, while the FakeSafe images $F(X)$ are different.  

\par 
In a similar manner, when conducting experiments on MNIST fashion data set which contain objects from 10 different categories, such as "shoe" or "dress", $R(F(X))$ have the same labels as the original message $X$ and could differ from labels of $X^{Fake}=F(X)$.  

\par
When conducting experiments on human face images, original data $X$ is a human face image which was mapped to another human image $X^{Fake}=F(X)$ that could be from the same person or a different person. 

When conducting experiments on English words, original data $X$ is a 50-dimension word embeddings which was mapped to another 300-dimension word embeddings $X^{Fake}=F(X)$ that could be from the same word or a different word. 

\begin{figure}
  \centering
  {\setlength{\fboxsep}{0pt}%
  \setlength{\fboxrule}{0pt}%
  \fbox{\includegraphics[width=8cm, height=8cm]{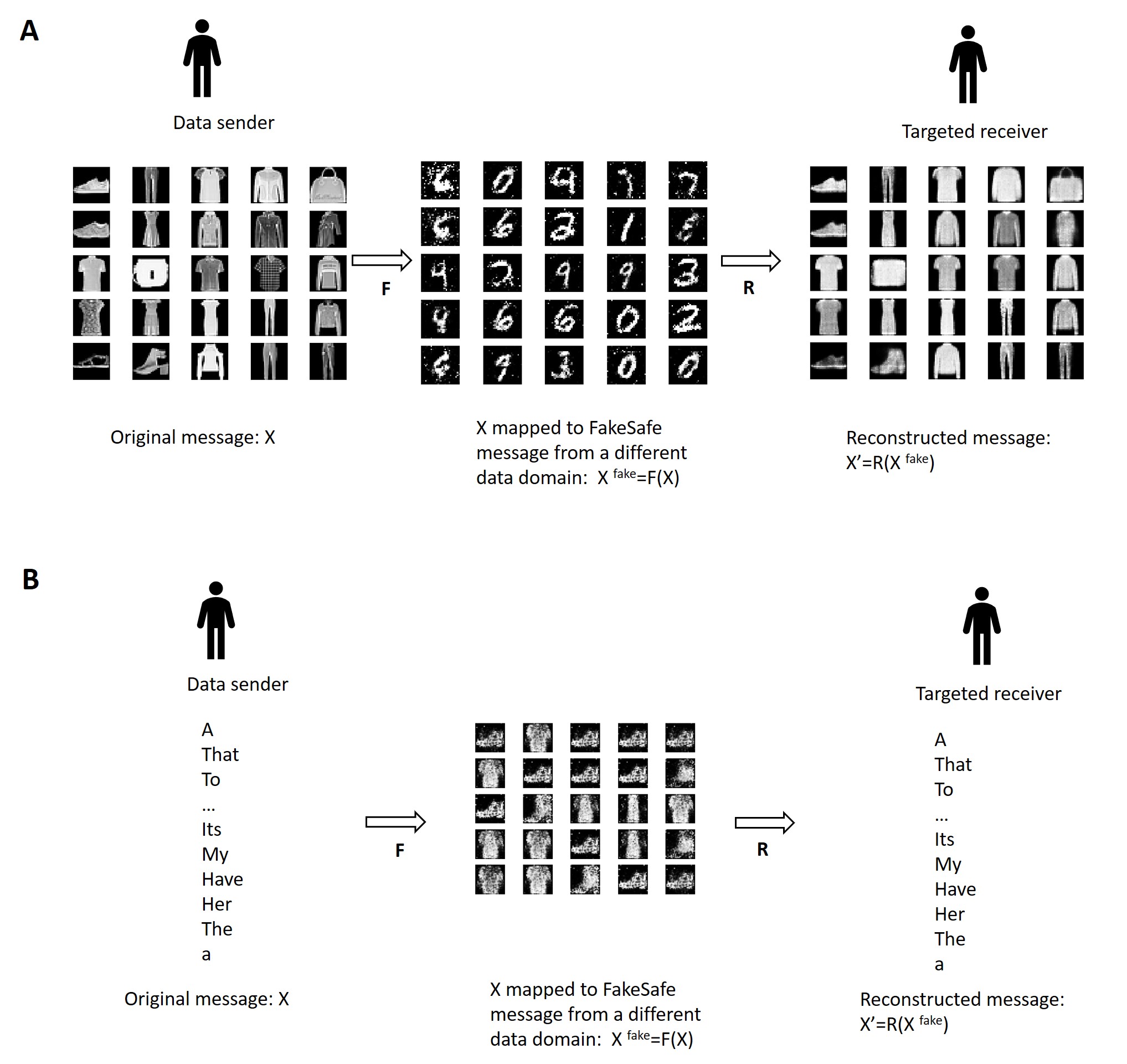}}
}%
 \label{CrossDomain}
  \caption{Use FakeSafe method to  map private information into  fake data in a different domain. (A)}
\end{figure}

\par 
In order to evaluate quality of the reconstructed message$R(F(X))$, two metrics were used. First, reconstruction errors between $R(F(X))$ and $X$ were calculated as mean squared errors. Second, in order to know whether the reconstructed messages $R(F(X))$ still look like from the same class or individual as $X$ to human, we trained a classifier $C$ on $X$ in the training set to classify their labels, i.e. the digits, fashion category or individual ID, and apply $C$  onto reconstructed data $R(F(X))$. The accuracy, F1 score, precision and recall of $C(R(F(X)))$ were compared with the original labels of $X$.

\begin{table*}
\caption{Performance of FakeSafe mapping onto the same domain or different domain}
\label{Performance}
\resizebox{17.5cm}{!}{%
\begin{tabular}{|c|c|c|c|c|c|c|c|c|}
\hline
\textbf{Experiment} & 
\textbf{Original message} & 
\textbf{FakeSafe message} &
\textbf{Reconstruction error} &
\textbf{Precision} &
\textbf{Recall} &
\textbf{F1 Score} &
\textbf{Type} &
\textbf{Remark} \\ 
\hline

\textbf{Face-\textgreater{}F-\textgreater{}Face-\textgreater{}R-\textgreater{}Face}          & Face image    & Face image    & 1.06 & 0.95 & 0.92 & 0.92 & Same domain&-  \\ \hline

\textbf{MNIST-\textgreater{}F-\textgreater{}MNIST-\textgreater{}R-\textgreater{}MNIST}       & MNIST digits  & MNIST digits  & 1.62 & 0.9  & 0.8  & 0.81 & Same domain&-  \\ \hline

\textbf{Fashion-\textgreater{}F-\textgreater{}Fashion-\textgreater{}R-\textgreater{}Fashion} & Fashion image & Fashion image & 1.2  & 0.71 & 0.72 & 0.7  & Same domain &- \\ \hline

\textbf{Word-\textgreater{}F-\textgreater{}Word-\textgreater{}R-\textgreater{}Word}       & English words (50d embeddings) & English words (300d embeddings) & NA   & 0.65  & 0.68 & 0.66 & Same domain & All 202 words \\ \hline

\textbf{Face-\textgreater{}F-\textgreater{}MNIST-\textgreater{}R-\textgreater{}Face}         & Face image    & MNIST digits  & 0.1  & 1    & 1    & 1    & Cross domain &- \\ \hline

\textbf{Fashion-\textgreater{}F-\textgreater{}MNIST-\textgreater{}R-\textgreater{}Fashion}   & Fashion image & MNIST digits  & 0.79 & 0.85 & 0.76 & 0.79 & Cross domain &- \\ \hline

\textbf{{MNIST-\textgreater{}F-\textgreater{}Fashion-\textgreater{}R-\textgreater{}MNIST}}       & MNIST digits & Fashion image & 1.32 & 0.91 & 0.88 & 0.88& Cross domain &- \\ \hline

\textbf{Word-\textgreater{}F-\textgreater{}Fashion-\textgreater{}R-\textgreater{}Word}       & English words (tokens) & Fashion image & NA   & 0.8  & 0.84 & 0.81 & Cross domain & Top 100 frequent words \\ \hline

\textbf{Word-\textgreater{}F-\textgreater{}Fashion-\textgreater{}R-\textgreater{}Word}       & English words (50d embeddings) & Fashion image & NA   & 0.96  & 0.96 & 0.96 & Cross domain & Top 100 frequent words \\ \hline

\textbf{Word-\textgreater{}F-\textgreater{}Fashion-\textgreater{}R-\textgreater{}Word}       & English words (50d embeddings) & Fashion image & NA   & 0.65  & 0.68 & 0.66 & Cross domain & All 202 words \\ \hline
\end{tabular}%
}
\end{table*}

\par
The MNIST$\rightarrow$F$\rightarrow$MNIST$\rightarrow$R$\rightarrow$MNIST FakeSafe experiment achieved a reconstruction error of 1.62, classifier precision of 0.90, recall of 0.80 and F1 score of 0.81. In the Fashion$\rightarrow$F$\rightarrow$Fashion$\rightarrow$R$\rightarrow$Fashion experiment (Table \ref{Performance} ), FakeSafe achieved a reconstruction error of 1.2, classifier precision of 0.71, recall of 0.72 and F1 score of 0.70. The real world human face image data set, FakeSafe method achieved a reconstruction error of 1.06, precision of 0.95, recall of 0.92 and F1 score of 0.92.

\subsection{FakeSafe mapping onto a different data domain}
Hide privacy information onto fake messages of the same type can help protect the information by misleading the malicious attackers. However, sometimes it is better not to expose the original information domain at all. Therefore, we conducted experiments to FakeSafe map information into the message in a different domain. We conducted 4 experiments, MNIST$\rightarrow$F$\rightarrow$Fashion$\rightarrow$R$\rightarrow$MNIST, Fashion$\rightarrow$F$\rightarrow$MNIST$\rightarrow$R$\rightarrow$Fashion, Face$\rightarrow$F$\rightarrow$MNIST$\rightarrow$R$\rightarrow$Face and Word$\rightarrow$F$\rightarrow$Fashion$\rightarrow$R$\rightarrow$Word. The performances are comparable to FakeSafe mapping onto the same data domain (Table \ref{Performance}). 
\par
Specifically, for the experiment Word$\rightarrow$F$\rightarrow$Fashion$\rightarrow$R$\rightarrow$Word, we have tried two different approaches to map the original messages. In the first approach, we will tokenize the original messages, which are the English words, and then map the tokens to MNIST fashion images using FakeSafe. During the decoding process, we will map the MNIST fashion images back to tokens, which will be eventually converted back to English words. In the second approach, we will first convert the words to word embeddings with 50 dimensions, using GloVe Word Embeddings, and then map the word embeddings to MNIST fashion images. During the decoding process, we will use FakeSafe to map the MNIST fashion images back to word embeddings, and then decode back to the original words by finding the word with the smallest cosine similarity with the decoded word embeddings. It is noteworthy that, the second approach, which uses word embeddings as the original messages, is proved to achieve better performance than the first approach, which only uses word tokens as the original messages.

\begin{figure*}
  \centering
  {\setlength{\fboxsep}{0pt}%
  \setlength{\fboxrule}{0pt}%
  \fbox{\includegraphics[width=16cm, height=4cm]{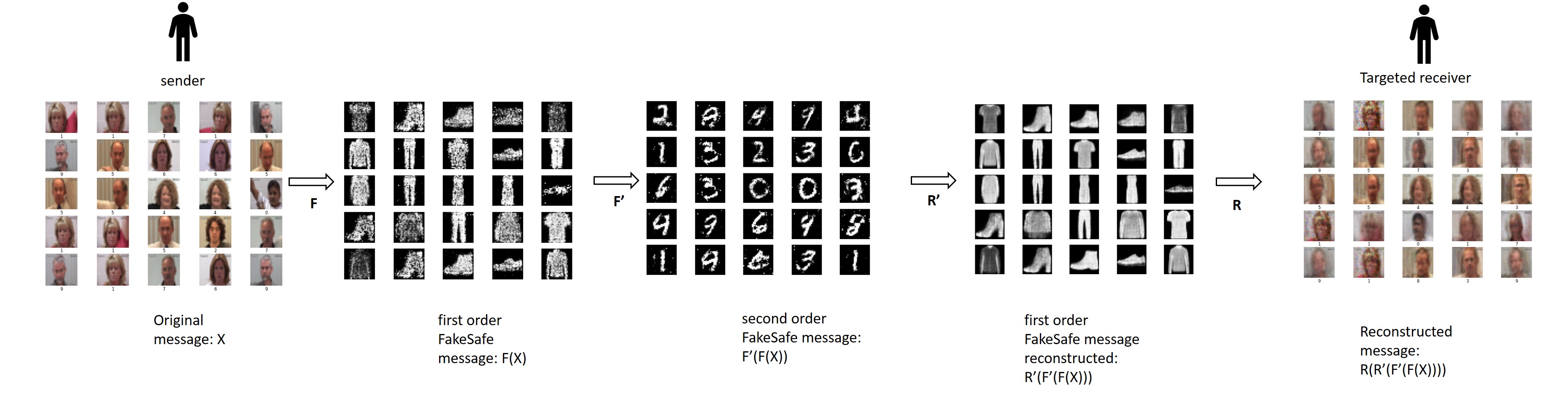}}
}%
 \label{Two-step}
  \caption{Multi-step FakeSafe mapping. The reconstruction errors increase with depth}
\end{figure*}

\begin{table}
 \caption{Performance of multi-step FakeSafe mapping)}
 \label{Multi-step-Performance}
 \resizebox{17.5cm}{!}{%
\begin{tabular}{|c|c|c|c|c|c|}
\hline
\textbf{Experiment} &
  \textbf{Reconstruction error} &
  \textbf{Precision} &
  \textbf{Recall} &
  \textbf{F1 Score} &
  \textbf{Type} \\ \hline
\textbf{mnist-\textgreater{}mnist-\textgreater{}fashion-\textgreater{}mnist-\textgreater{}mnist} &
  9.4 &
  0.75 &
  0.64 &
  0.63 &
  two-step \\ \hline
\textbf{fashion-\textgreater{}mnist-\textgreater{}mnist-\textgreater{}mnist-\textgreater{}fashion} &
  1.1 &
  0.81 &
  0.72 &
  0.73 &
  two-step \\ \hline
\textbf{face-\textgreater{}fashion-\textgreater{}mnist-\textgreater{}fashion-\textgreater{}face} &
  37.1 &
  0.66 &
  0.56 &
  0.59 &
  two-step \\ \hline
\textbf{mnist-\textgreater{}mnist-\textgreater{}fashion-\textgreater{}fashion-\textgreater{}fashion-\textgreater{}mnist-\textgreater{}mnist} &
  17.3 &
  0.48 &
  0.44 &
  0.39 &
  three-step \\ \hline
\textbf{face-\textgreater{}mnist-\textgreater{}fashion-\textgreater{}fashion-\textgreater{}fashion-\textgreater{}mnist-\textgreater{}face} &
  88.8 &
  0.29 &
  0.36 &
  0.29 &
  three-step \\ \hline
\end{tabular}
}
\end{table}

\section{Deeper FakeSafe mapping}
To guarantee the safety of sensitive data, one may ask why not map the original private multiple times using a cascade of different $F$ functions, so that even if the attacker knows the message is fake, he or she will not know how many steps the messages were mapped. In order to explore the feasibility of deeper FakeSafe mapping, we conducted a series of experiments of 2-step and 3-step FakeSafe using MNIST, fashion and face images (Figure \ref{Two-step} and Table \ref{Multi-step-Performance}). Our results suggest that even it is possible to conduct multi-step FakeSafe mapping, the reconstruction error increased and classification accuracy decreased dramatically.

\clearpage

\section{Conclusion}
In this article, we propose a method, named FakeSafe, to provide human-level private data protection by mapping each data point into a fake message that looks realistic to human. We utilized GANs with cycle-consistency to build a function to map the original data to fake message and another function to map the fake message back to the original data. Both functions are data set specific and can be easily adjusted for the other data sets. FakeSafe method gives users flexibility to map private data onto different data domains depending on use cases. In addition, FakeSafe can be easily used in combination with traditional data protection technologies but focus on human-level protection which takes human factors in data security and privacy into consideration.

\bibliographystyle{aaai}\bibliography{neurips_2020}

\begin{thebibliography}{}

\bibitem[\protect\citeauthoryear{Bengio and others}{2009}]{bengio2009learning}
Bengio, Y., et~al.
\newblock 2009.
\newblock Learning deep architectures for ai.
\newblock {\em Foundations and trends{\textregistered} in Machine Learning}
  2(1):1--127.

\bibitem[\protect\citeauthoryear{Goodfellow \bgroup et al\mbox.\egroup
  }{2014}]{goodfellow2014generative}
Goodfellow, I.; Pouget-Abadie, J.; Mirza, M.; Xu, B.; Warde-Farley, D.; Ozair,
  S.; Courville, A.; and Bengio, Y.
\newblock 2014.
\newblock Generative adversarial nets.
\newblock In {\em Advances in neural information processing systems},
  2672--2680.

\bibitem[\protect\citeauthoryear{Jensen}{2013}]{jensen2013challenges}
Jensen, M.
\newblock 2013.
\newblock Challenges of privacy protection in big data analytics.
\newblock In {\em 2013 IEEE International Congress on Big Data},  235--238.
\newblock IEEE.

\bibitem[\protect\citeauthoryear{Joly \bgroup et al\mbox.\egroup
  }{2016}]{joly2016data}
Joly, Y.; Dyke, S.~O.; Knoppers, B.~M.; and Pastinen, T.
\newblock 2016.
\newblock Are data sharing and privacy protection mutually exclusive?
\newblock {\em Cell} 167(5):1150--1154.

\bibitem[\protect\citeauthoryear{Jordon, Yoon, and van~der
  Schaar}{2018}]{jordon2018pate}
Jordon, J.; Yoon, J.; and van~der Schaar, M.
\newblock 2018.
\newblock Pate-gan: Generating synthetic data with differential privacy
  guarantees.

\bibitem[\protect\citeauthoryear{Konečný \bgroup et al\mbox.\egroup
  }{2017}]{godard2017unsupervised}
Konečný, J.; McMahan, B.; Yu, F.; Richtárik, P.; Suresh, A.~T.; and Bacon,
  D.
\newblock 2017.
\newblock Unsupervised monocular depth estimation with left-right consistency.
\newblock In {\em Proceedings of the IEEE Conference on Computer Vision and
  Pattern Recognition},  270--279.

\bibitem[\protect\citeauthoryear{Lee and Gostin}{2009}]{lee2009ethical}
Lee, L.~M., and Gostin, L.~O.
\newblock 2009.
\newblock Ethical collection, storage, and use of public health data: a
  proposal for a national privacy protection.
\newblock {\em Jama} 302(1):82--84.

\bibitem[\protect\citeauthoryear{Liu, Dligach, and Miller}{2019}]{liu2019two}
Liu, D.; Dligach, D.; and Miller, T.
\newblock 2019.
\newblock Two-stage federated phenotyping and patient representation learning.
\newblock In {\em Proceedings of the 18th BioNLP Workshop and Shared Task},
  283--291.

\bibitem[\protect\citeauthoryear{Lucey \bgroup et al\mbox.\egroup
  }{2011}]{lucey2011painful}
Lucey, P.; Cohn, J.~F.; Prkachin, K.~M.; Solomon, P.~E.; and Matthews, I.
\newblock 2011.
\newblock Painful data: The unbc-mcmaster shoulder pain expression archive
  database.
\newblock In {\em Face and Gesture 2011},  57--64.
\newblock IEEE.

\bibitem[\protect\citeauthoryear{Makhzani \bgroup et al\mbox.\egroup
  }{2015}]{makhzani2015adversarial}
Makhzani, A.; Shlens, J.; Jaitly, N.; Goodfellow, I.; and Frey, B.
\newblock 2015.
\newblock Adversarial autoencoders.
\newblock {\em arXiv preprint arXiv:1511.05644}.

\bibitem[\protect\citeauthoryear{Mirza and
  Osindero}{2014}]{mirza2014conditional}
Mirza, M., and Osindero, S.
\newblock 2014.
\newblock Conditional generative adversarial nets.
\newblock {\em arXiv preprint arXiv:1411.1784}.

\bibitem[\protect\citeauthoryear{Mukherjee \bgroup et al\mbox.\egroup
  }{2019}]{mukherjee2019protecting}
Mukherjee, S.; Xu, Y.; Trivedi, A.; and Ferres, J.~L.
\newblock 2019.
\newblock Protecting gans against privacy attacks by preventing overfitting.
\newblock {\em arXiv preprint arXiv:2001.00071}.

\bibitem[\protect\citeauthoryear{Odena, Olah, and
  Shlens}{2017}]{odena2017conditional}
Odena, A.; Olah, C.; and Shlens, J.
\newblock 2017.
\newblock Conditional image synthesis with auxiliary classifier gans.
\newblock In {\em Proceedings of the 34th International Conference on Machine
  Learning-Volume 70},  2642--2651.
\newblock JMLR. org.

\bibitem[\protect\citeauthoryear{Radford, Metz, and
  Chintala}{2015}]{radford2015unsupervised}
Radford, A.; Metz, L.; and Chintala, S.
\newblock 2015.
\newblock Unsupervised representation learning with deep convolutional
  generative adversarial networks.
\newblock {\em arXiv preprint arXiv:1511.06434}.

\bibitem[\protect\citeauthoryear{Shao, Liu, and Liu}{2019}]{shao2019privacy}
Shao, R.; Liu, H.; and Liu, D.
\newblock 2019.
\newblock Privacy preserving stochastic channel-based federated learning with
  neural network pruning.
\newblock {\em arXiv preprint arXiv:1910.02115}.

\bibitem[\protect\citeauthoryear{Zhou \bgroup et al\mbox.\egroup
  }{2016}]{zhou2016learning}
Zhou, T.; Krahenbuhl, P.; Aubry, M.; Huang, Q.; and Efros, A.~A.
\newblock 2016.
\newblock Learning dense correspondence via 3d-guided cycle consistency.
\newblock In {\em Proceedings of the IEEE Conference on Computer Vision and
  Pattern Recognition},  117--126.

\bibitem[\protect\citeauthoryear{Zhu \bgroup et al\mbox.\egroup
  }{2017}]{zhu2017unpaired}
Zhu, J.-Y.; Park, T.; Isola, P.; and Efros, A.~A.
\newblock 2017.
\newblock Unpaired image-to-image translation using cycle-consistent
  adversarial networks.
\newblock In {\em Proceedings of the IEEE international conference on computer
  vision},  2223--2232.

\end{thebibliography}

\clearpage

\section{Appendix}

\par
Furthermore, we have also conducted a supplementary FakeSafe experiment case Sentence$\rightarrow$F$\rightarrow$Fashion$\rightarrow$R$\rightarrow$Sentence, the performance of which is demonstrated in Table \ref{Performance-Supplementary}. We have tried two different approaches to conduct the experiment. 

\par
In the first approach, we have trained a Seq2Seq model using the GRU layer, which will encode the sentence sequence to the internal hidden states, and then decode back to the original sentence sequence. Then we will map the internal states, which are generated by the Seq2Seq encoder, to MNIST fashion images. During the decoding process, we will use FakeSafe to map the MNIST fashion images back to the internal states used by the Seq2Seq model, and further decode back to the sentence. 

\par
In the second approach, considering that Word$\rightarrow$F$\rightarrow$Fashion$\rightarrow$R$\rightarrow$Word model achieves good performance on 50-dimension words embeddings, we have attempted to train a Word$\rightarrow$F$\rightarrow$Fashion$\rightarrow$R$\rightarrow$Word model first. Then we have split the sentence into a list of words, and each of the words will be encoded into a MNIST fashion image. Eventually, the MNIST fashion images will be decoded back to the list of words, which will be further converted to the sentences.

\begin{table}
\caption{Performance of Supplementary Experiments}
\label{Performance-Supplementary}
\resizebox{17.5cm}{!}{%
\begin{tabular}{|c|c|c|c|c|c|c|}
\hline
\textbf{Experiment} &
  \textbf{Original message} &
  \textbf{FakeSafe message} &
  \textbf{BLEU score} &
  \textbf{Type} &
  \textbf{Data set} &
  \textbf{Remark} \\ \hline
\textbf{Sentence-\textgreater{}F-\textgreater{}Fashion-\textgreater{}R-\textgreater{}Sentence}       & Sentence states & Fashion image & 0.04 & Cross domain & Tatoeba (7761 words) & Use seq2seq model \\ \hline
\textbf{Sentence-\textgreater{}F-\textgreater{}Fashion-\textgreater{}R-\textgreater{}Sentence}       & Sentence states & Fashion image & 0.33 & Cross domain & {small}\textunderscore{vocab}\textunderscore{en} (202 words) & Use seq2seq model \\ \hline
\textbf{Sentence-\textgreater{}F-\textgreater{}Fashion-\textgreater{}R-\textgreater{}Sentence}       & English words (50d embeddings) & Fashion image & 0.48 & Cross domain & {small}\textunderscore{vocab}\textunderscore{en} (202 words) & Use word embeddings \\ \hline
\end{tabular}%
}
\end{table}

\end{document}